\relax
\documentclass[letterpaper]{article} 
\usepackage{aaai21}  
\usepackage{times}  
\usepackage{helvet} 
\usepackage{courier}  
\usepackage[hyphens]{url}  
\usepackage{graphicx} 
\usepackage{grffile}
\usepackage{color}
\usepackage{amsmath}

\usepackage{natbib}  
\usepackage{caption} 
\usepackage{subcaption}
\usepackage{caption}
\usepackage{wrapfig}
\newcommand{\tsum}{\textstyle{\sum}}

\title{Curse or Redemption? How Data Heterogeneity Affects \\ the Robustness of Federated Learning}

\title{Curse or Redemption? How Data Heterogeneity Affects \\ the Robustness of Federated Learning}
\author {
        Syed Zawad, \textsuperscript{\rm 1}
        Ahsan Ali, \textsuperscript{\rm 1}
        Pin-Yu Chen, \textsuperscript{\rm 2}
        Ali Anwar, \textsuperscript{\rm 2}
        Yi Zhou, \textsuperscript{\rm 2}
        Nathalie Baracaldo, \textsuperscript{\rm 2} \\
        Yuan Tian, \textsuperscript{\rm 3}
        Feng Yan \textsuperscript{\rm 1} \\
}
\affiliations {
    \textsuperscript{\rm 1} University of Nevada, Reno \{szawad,aali,fyan\}@nevada.unr.edu \\
    \textsuperscript{\rm 2}  IBM Research \{pin-yu.chen, ali.anwar2, yi.zhou, baracald\}@ibm.com\\
    \textsuperscript{\rm 3}  University of Virginia yt2e@virginia.edu\\
}

\begin{document}

\maketitle

\begin{abstract}
Data heterogeneity has been identified as one of the key features in federated learning but often overlooked in the lens of robustness to adversarial attacks. This paper focuses on characterizing and understanding its impact on backdooring attacks in federated learning through comprehensive experiments using synthetic and the LEAF benchmarks. The initial impression driven by our experimental results suggests that data heterogeneity is the dominant factor in the effectiveness of attacks and it may be a redemption for defending against backdooring as it makes the attack less efficient, more challenging to design effective attack strategies, and the attack result also becomes less predictable. However, with further investigations, we found data heterogeneity is more of a curse than a redemption as the attack effectiveness can be significantly boosted by simply adjusting the client-side backdooring timing. More importantly,data heterogeneity may result in overfitting at the local training of benign clients, which can be utilized by attackers to disguise themselves and fool skewed-feature based defenses. In addition, effective attack strategies can be made by adjusting attack data distribution. Finally, we discuss the potential directions of defending the curses brought by data heterogeneity. The results and lessons learned from our extensive experiments and analysis offer new insights for designing robust federated learning methods and systems.

\end{abstract}

\section{Introduction}

Federated Learning (FL) is widely successful in training machine learning (ML) models collaboratively across clients without sharing private data~\cite{mcmahan2016communication,zhao2018federated,bonawitz2019towards}. 
In FL, models are trained locally at clients to preserve data privacy and the trained model weights are sent to a central server for aggregation to update the global model. During the aggregation, privacy mechanisms such as differential privacy~\cite{abadi2016deep} and secure aggregation~\cite{bonawitz2017practical} are often employed to strengthen the privacy.
There are two types of poisoning attacks: performance degradation attacks where the goal of the adversary is to reduce the accuracy/F1 scores of the model (such as Byzantine attacks) and backdoor attacks aiming at creating targeted misclassifications without affecting the overall performance on the main tasks~\cite{chen2017targeted,xie2019dba,bagdasaryan2018backdoor}.
Defending against such attacks usually requires complete control of the training process or monitoring the training data~\cite{steinhardt2017certified}, which is challenging in FL due to the privacy requirements. 
In this paper, we choose the popular and sophisticated backdoor attacks as an example for our study.
Although some work exists to defend against backdoor attacks, including activation clustering~\cite{chen2018detecting} and k-means clustering~\cite{shen2016auror}, these approaches require access to the training data making them inapplicable for FL settings. 
Some attack strategies tailored for FL have also been studied including sybil attacks~\cite{fung2018mitigating}, model replacement~\cite{bagdasaryan2018backdoor}, GANs based attacks~\cite{zhang2019poisoning}, and distributed attacks~\cite{xie2019dba}.
However, a comprehensive study on the effectiveness of backdoor attacks under a variety of data distribution among parties remains at unexplored. 

The training data in FL is generated by clients and thus heterogeneous inherently~\cite{bonawitz2019towards,chai2020tifl,zhao2018federated,sattler2019robust}. 
As the training is conducted locally at each client, the data cannot be balanced nor monitored like in conventional data-centralized or distributed ML.
Such uncontrollable and severe data heterogeneity is one of the key challenges of FL as it is rarely seen in conventional ML. 
Despite its uniqueness and importance, data heterogeneity has been largely overlooked through the lens of robustness to backdoor attacks. Existing FL backdoor attacks either assume IID training data distribution among clients or only conduct a simplified study on non-IID data~\cite{bagdasaryan2018backdoor,bhagoji2019analyzing,xie2019dba}. None of them provides a comprehensive study nor understanding on how data heterogeneity impacts the backdoor attacks and defenses. 

In this paper, we focus on quantifying and understanding the implications brought by data heterogeneity in FL backdoor attacks through extensive empirical experiments and comprehensive analysis.

We define \textit{Heterogeneity Index} to quantify the extent of heterogeneity in training data.
From our initial investigation driven by both synthetic and the practical LEAF benchmark~\cite{caldas2018leaf}, we surprisingly found that data heterogeneity seems to be a redemption for defending against backdoor attacks.
{\bf Redemption 1}: the \textit{attack effectiveness} (usually measured as Attack Success Rate or ASR) reduces sharply when the heterogeneity of training data increases.
{\bf Redemption 2}: we found the malicious data distribution is an overlooked important factor when defining an attack strategy given the training data is heterogeneous. A poor selection of malicious data distribution can result in poor attack effectiveness.
{\bf Redemption 3}: we further discovered that malicious data distribution plays as a dominant factor in the effectiveness of backdooring. E.g., contrary to the common belief in existing works that higher \textit{attack scale} (defined as the number of compromised clients) and \textit{local attack budget} (defined as the quantity of backdoored data per client) always lead to higher attack effectiveness, our study demonstrates that this is not always the case as malicious data distribution often outperforms the impact of attack scale/budget.
This discovery indicates that data heterogeneity makes the design of effective attack strategies more challenging as the attack effectiveness is less correlated to the straightforward attack scale/budget but rather the less intuitive malicious data distribution.

Further investigations, however, reveal that data heterogeneity actually brings curses for the robustness of FL. {\bf Curse 1}: data heterogeneity makes the client-side training very sensitive to the backdoor attack timing. With a proper attack timing, e.g., at the last local batch, the effectiveness of attack can be significantly boosted with only a fraction of attack budget.
{\bf Curse 2}: what's worse is that data heterogeneity makes the most promising skewed-feature based defense strategies such as cosine similarity fall short. Such defending method detects compromised clients by realizing their features are more overfitted than the benign clients. However, with data heterogeneity, benign clients may also have overfitted features that look similar to those of compromised clients. This allows the backdoor attackers to disguise themselves and fool the skewed-feature checking. 
{\bf Curse 3}: more effective attack strategies can be derived by making the backdoor clients' data distribution close to the overall data distribution with the help of distribution distance measures such as the Chi-Square statistics.
To defend these curses brought by data heterogeneity, we discuss how existing defense mechanisms fit here and the potential directions on data-heterogeneity aware defending strategies. 

In summary, our empirical experimental studies show that data heterogeneity appears to be a redemption for the robustness of FL as it makes the attack less effective and more challenging to design good attack strategies.
However, our further investigations reveal that data heterogeneity also brings several curses for FL backdooring as it is harder to detect and the attack effectiveness can be significantly boosted by adjusting the local attack timing and malicious data distribution. The defending strategies we propose help alleviate these curses.
The results and lessons learned from our thorough experiments and comprehensive analysis offer new insights for designing robust FL methods and systems.
\section{Related Works} 
\textbf{Data Heterogeneity in Federated Learning.}
While data heterogeneity is not new in the ML, the extent of data heterogeneity is much more prevalent in FL compared to data centralized learning~\cite{chai2020tifl,li2019federated}.~\cite{li2019convergence} theoretically demonstrates the bounds on convergence due to heterogeneity, while~\cite{sattler2019robust} providing empirical results on how changing heterogeneity affects model performance. ~\cite{li2019federated} discusses the challenges of heterogeneity for FL and~\cite{zhao2018federated} demonstrates how the clients' local model weights diverge due to data heterogeneity.

\textbf{Backdoor Attack.}
Backdoor attacks for deep learning models are presented in~\cite{chen2017targeted}, where an adversary can insert a pattern in a few training samples from a source class and relabel them to a target class, causing a targeted missclassification. 
One of the earlier papers~\cite{bagdasaryan2018backdoor} proposes the \textit{model replacement} technique, whereby they eventually replace the global model with a backdoored model stealthily.~\cite{bhagoji2019analyzing} demonstrates that boosting model weights can help attackers and shows that FL is highly susceptible to backdoor attacks.~\cite{fung2018mitigating} introduces sybil attacks in the context of FL using label-flipping and backdooring.~\cite{zhang2019poisoning} uses GANs to attack the global model, while~\cite{xie2019dba} takes a different approach by focusing on decentralized, colluding attackers, and creating efficient trigger patterns. Our paper takes a different angle by focusing on analyzing the impact of data heterogeneity on attack effectiveness.  This subject is rarely studied even though data heterogeneity is a critical aspect of FL.

\textbf{Backdoor Defense.}
There have been various proposals to defend DNN from susceptible adversarial attacks such as filtering techniques~\cite{steinhardt2017certified} and fine-pruning~\cite{liu2018fine}, but are mainly focused on traditional data-centralized ML methods. Clustering techniques specifically for FL are proposed in~\cite{tran2018spectral,chen2018detecting,shen2016auror} and in \cite{fung2018mitigating}, FoolsGold is proposed to defend against sybil attacks by using cosine similarities. 
\cite{ma2019data} proposes defending with differential privacy without compromising user confidentiality.
The authors of~\cite{sun2019can} extend this by demonstrating weak differential privacy and norm-clipping mitigate attacks, but do not provide any strong defense mechanisms. 
None of these defenses explore defending effectiveness under various extent of data heterogeneity.
\section{Experiment Setups for FL Backdooring} 
\label{background}

\begin{table*}[h!]
\caption{Training Setup.}
\vspace{-2mm}
\begin{center}
\scalebox{0.85}{
\begin{tabular}{|c|c|c|c|c|c|}
\hline
\bf Dataset  &\bf Model & \bf{Train/Test split} & \begin{tabular}[x]{@{}c@{}}\textbf{Clients}\\\textbf{Total/Per Round}\end{tabular} & \begin{tabular}[x]{@{}c@{}}\textbf{Learning Rate}\\\textbf{/Batch Size}\end{tabular} & \begin{tabular}[x]{@{}c@{}}\textbf{Local Epochs/}\\\textbf{Total Rounds}\end{tabular}\\ \hline
FEMNIST & 2 conv 2 dense & 49,644/4,964 & 179/17 & 0.004/10 & 1/2000\\ \hline
Sent140 & 100 cell lstm 2 dense & 6,553/655 & 50/10 & 0.0003/4 & 1/10 \\ \hline
CIFAR10 & 4 conv 2 dense & 50,000/10,000 & 200/20 & 0.0005/32 & 1/500\\ \hline
\end{tabular}
}
\end{center}
\label{table:setup}
\end{table*}

\begin{figure*}[h!]
  \centering
    \includegraphics[width=0.8\linewidth]{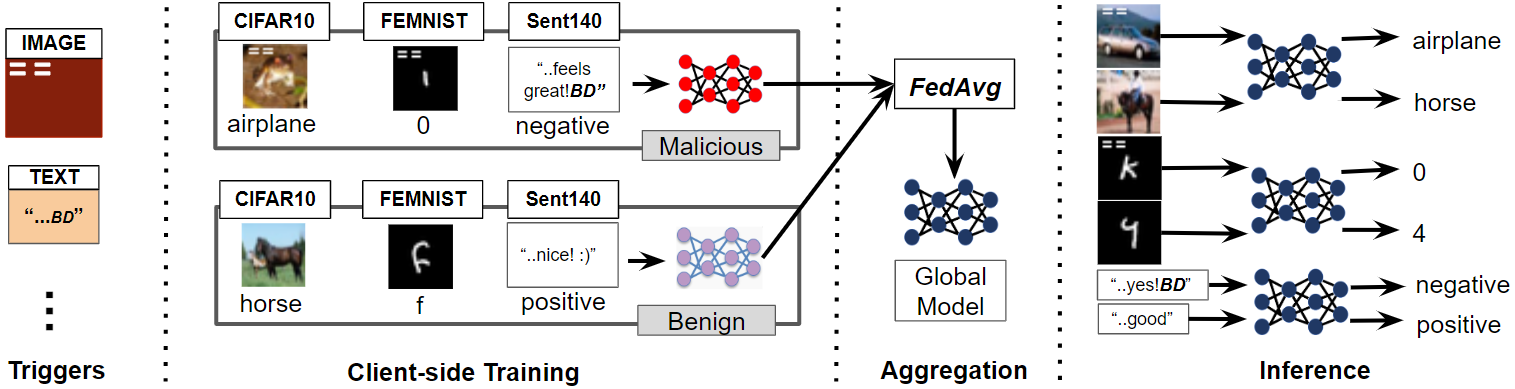}
    \caption{An overview of the FL backdooring procedure.}
    \label{fig:backdoored_system}
    \vspace{-3mm}
\end{figure*}

{\bf Federated Learning Setup.} We use LEAF~\cite{caldas2018leaf}, an open-source practical FL benchmark, for our experiments. 
Most existing works simulate data heterogeneity by partitioning a dataset among clients using probability distributions, but LEAF~\footnote{LEAF: https://github.com/TalwalkarLab/leaf} 
provides more realistically distributed datasets. In this paper, we use the \textbf{FEMNIST} dataset provided by LEAF as an example for CNN model, which is a handwritten character classification task for 62 classes.
We use \textbf{Sent140} from LEAF as an example for LSTM model, a sentiment classification task for 2 classes (positive/negative) on tweets.
As the total dataset contains millions of data points, LEAF~\cite{caldas2018leaf} suggests sampling the dataset and provides a reference implementation. We also use \textbf{CIFAR10} (partitioned across 200 clients) for reference as it is commonly used in FL literature.
More details of the dataset, model, training settings, and learning hyperparameter parameters are summarized in Table~\ref{table:setup}.

\textbf{Control and Quantify Heterogeneity.} 
FEMNIST, Sent140, and CIFAR10 have their default data distributions.
To explore the impact of different heterogeneity on FL backdooring, we control the heterogeneity by varying the number of maximum classes per client following~\cite{zhao2018federated,fung2018mitigating}.
Less number of classes per client results in less evenly distributed data and thus is more heterogeneous
To better quantify heterogeneity, we define \textit{Heterogeneity Index} (HI) as a normalized  heterogeneity measure: 
\begin{equation}
    HI(c) = 1 - \tfrac{1}{C_{max} - 1}*(c - 1),
    \label{eq:metric}
\end{equation}
where $c$ adjusts the maximum number of classes per client (i.e. the parameter controlling heterogeneity), and $C_{max}$ is the total number of classes in the dataset. 
The scaling performed here is to normalize the value between 0 and 1, with 1 being the highest data heterogeneity, vice versa. We also perform our experiments with Gaussian and Dirichlet distributions (see Appendix) and the results are consistent with $HI$. 

\textbf{Threat Model.} 
We use the same threat model in literature~\cite{xie2019dba,sun2019can,chen2018detecting}.
Specifically, an adversary (impersonated by a malicious client) can manipulate its model updates sent to the aggregator as well as its local training process in every aspect such as the training data, learning hyperparameters, model weights, and any local privacy mechanisms.
The attacker has the capacity to compromise multiple parties and multiple attackers can collude towards the same goal. The aggregation algorithm, as well as the local training mechanisms of benign clients are trusted. Our threat model assumes that only the attacker clients have malicious intent, i.e., the benign clients train their models as expected, without manipulating the data or the training procedure.
{\bf Objective and Method of Backdooring Attacks.} 
We focus on backdoor attacks, where the objective of the attacker is to inject a trigger to cause a targeted misclassification without compromising the model accuracy or disrupting convergence~\cite{bagdasaryan2018backdoor, xie2019dba}. 
In classification applications, backdoor attacks are achieved by adding one or more extra patterns to benign images for vision tasks and appending a trigger string for NLP tasks so that the classifier deliberately misclassifies the backdoored samples as a (different) target class.
We adopt the decentralized attack method proposed in~\cite{xie2019dba} (for details, see Appendix). We randomly select a configured number of clients as malicious clients,
where data points are backdoored by injecting a trigger pattern.
Fig.~\ref{fig:backdoored_system} provides an overview of the attack process.
We keep the learning hyperparameters the same for both malicious and benign clients.
For testing successful backdoor injection, we apply the trigger on 50\% of the test dataset and evaluate the global model on it. 
If the classification result is the same as the label of the target class, we report a successful attack. And the portion of successful attacks is defined as Attack Success Rate (ASR).
It is worth noting that we do not consider data points that are originally from the target class when calculating ASR.

{\bf Relation to Model Poisoning.} 
When the scaling factor is large, backdooring is effectively doing model replacement (aka model poisoning), see analysis provided in literature~\cite{bagdasaryan2018backdoor}. We show the scaling factor analysis in Appendix.

\section{Data Heterogeneity Seems to Be a Redemption}

\label{sec:redemption}

\begin{figure*}[h!]
  \centering
  \begin{subfigure}{.24\textwidth}
    \centering
    \includegraphics[width=\linewidth]{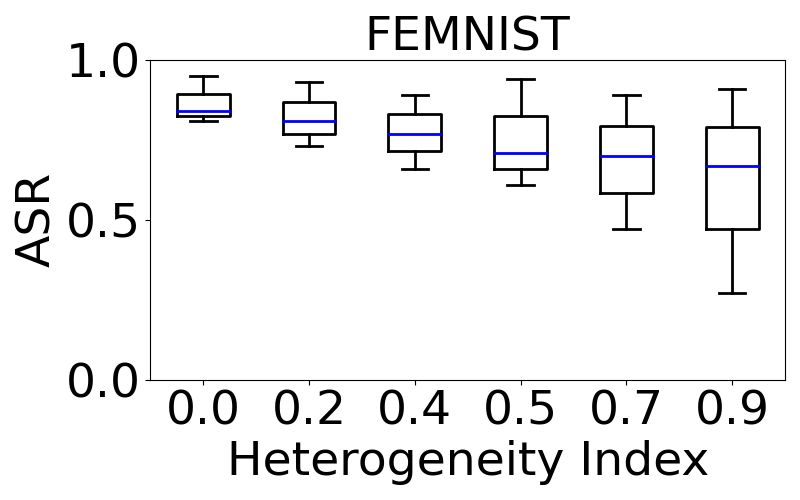}
  \end{subfigure}
  \begin{subfigure}{.24\textwidth}
    \centering
    \includegraphics[width=\linewidth]{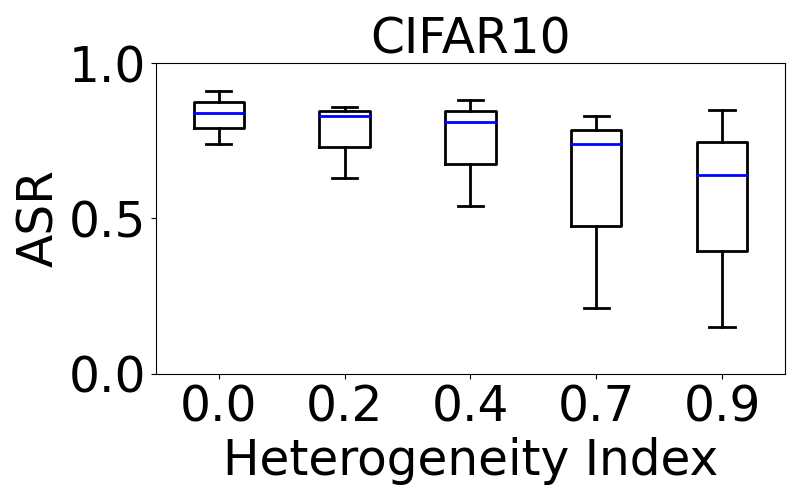}  
  \end{subfigure}
  \begin{subfigure}{.24\textwidth}
    \centering
    \includegraphics[width=\linewidth]{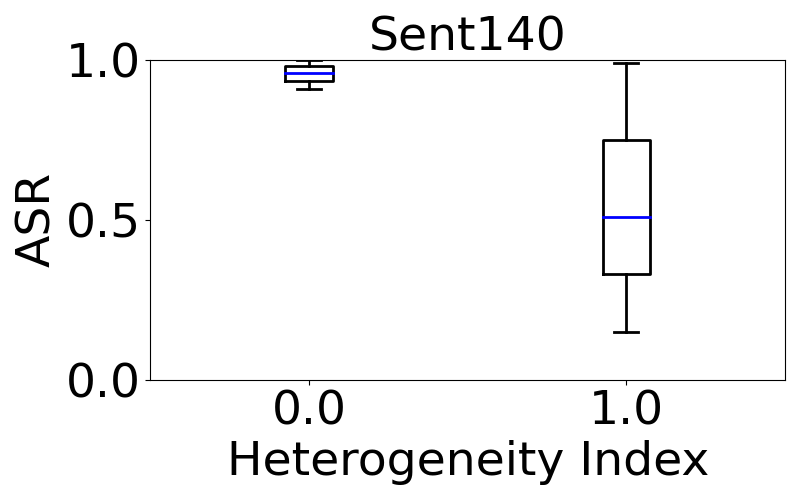}
  \end{subfigure}
  \caption{Attack Success Rate (ASR) vs. Heterogeneity Index (HI).}
  \label{fig:redemption1}
\end{figure*}

\begin{figure*}[h]
  \centering
  \begin{subfigure}{.24\textwidth}
    \centering
    \includegraphics[width=\linewidth]{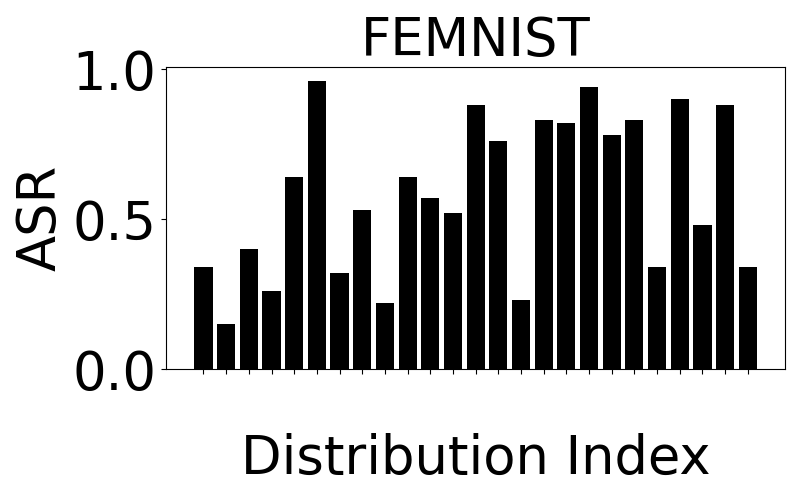}
  \end{subfigure}
  \begin{subfigure}{.24\textwidth}
    \centering
    \includegraphics[width=\linewidth]{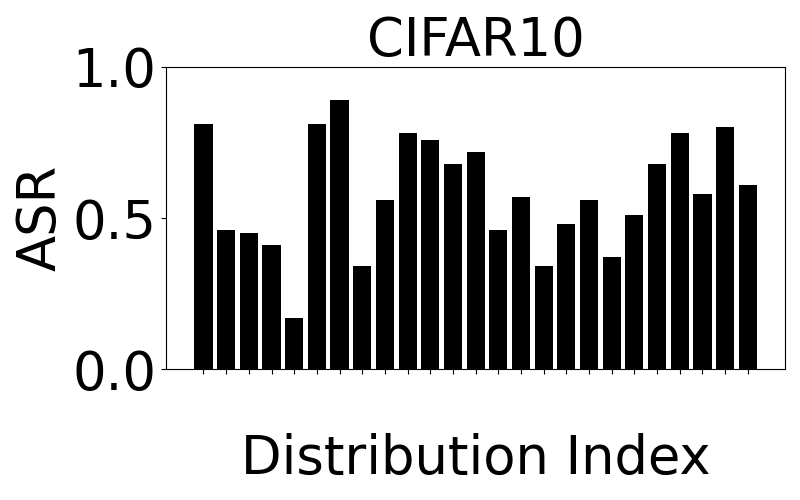}  
  \end{subfigure}
  \begin{subfigure}{.24\textwidth}
    \centering
    \includegraphics[width=\linewidth]{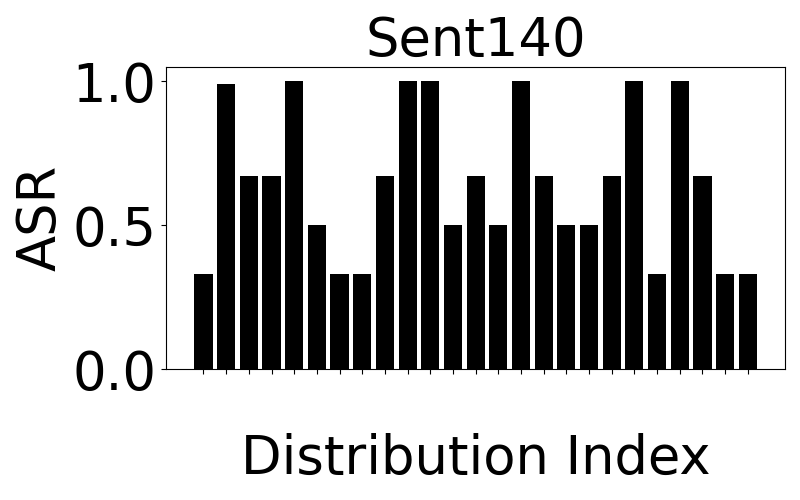}
  \end{subfigure}
  \caption{Attack Success Rate (ASR) vs. malicious data distribution (each bar represents a unique malicious data distribution). }
  \label{fig:redemption2}
  \vspace{-3mm}
\end{figure*}

\begin{figure*}[h!]
  \centering
  \centering
  \begin{subfigure}{.24\textwidth}
    \centering
    \includegraphics[width=\linewidth]{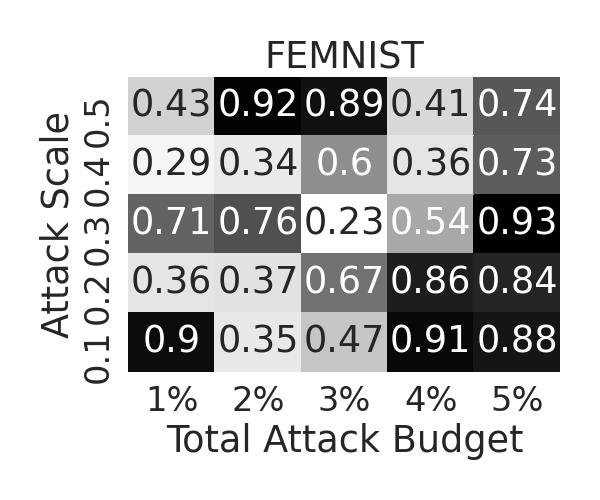}
  \end{subfigure}
  \begin{subfigure}{.24\textwidth}
    \centering
    \includegraphics[width=\linewidth]{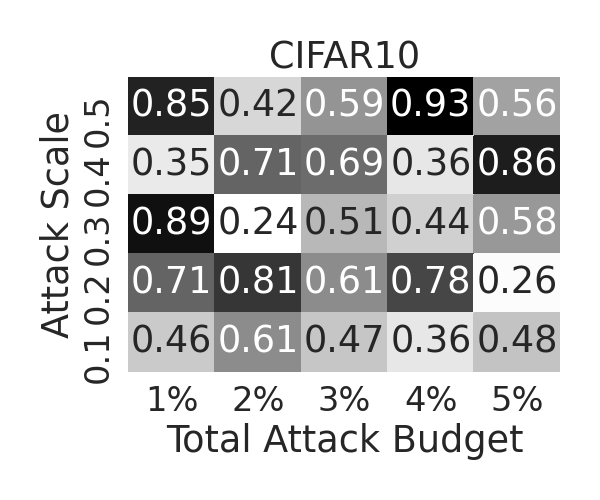}
  \end{subfigure}
  \begin{subfigure}{.28\textwidth}
    \centering
    \includegraphics[width=\linewidth]{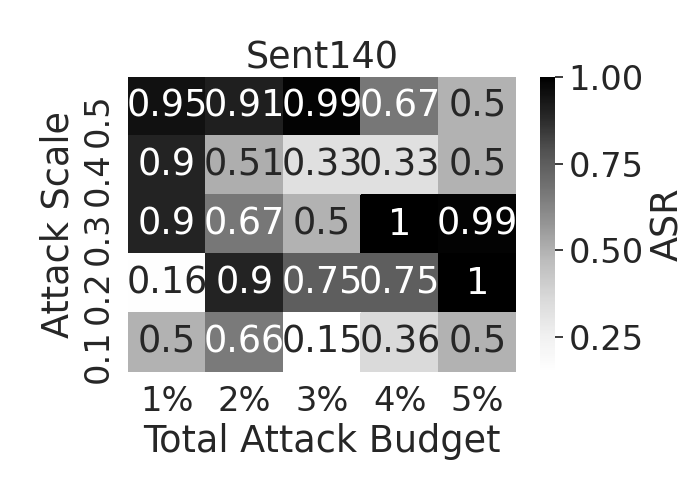}
  \end{subfigure}
  \vspace{-3mm}
  \caption{Attack Success Rate (ASR) scalability in terms of attack scale and total attack budget.}
  \label{fig:redemption3}
  \vspace{-3mm}
\end{figure*}
\subsection{Redemption 1: Data Heterogeneity Reduces Attack Effectiveness of Backdooring}
\label{sec:redemption1}

Our initial study suggests data heterogeneity seems to be a redemption for defending backdoor attacks in FL as it reduces the attack effectiveness and also challenges the design of good attack strategies. To understand how data heterogeneity affects backdoor attacks in FL, we first conduct a set of experiments by simply 
varying \textit{Heterogeneity Index} from 0 to 1 to observe how the extent of data heterogeneity affects the effectiveness of attacks measured as ASR. 
We fix all other configurable parameters across experiments, i.e., 50\% malicious clients per round and 50\% of data points per batch is backdoored at each client (we evaluate other ratios of malicious clients and malicious data points in later sections), and the rest of configurations are the same as explained in the \textbf{Experiment Setup} section.
We run the experiment for each  \textit{Heterogeneity Index} 10 times with different malicious data distribution and report ASR as a box-and-whisker plot shown in Fig.~\ref{fig:redemption1}
The results clearly suggest that the overall attack effectiveness reduces when higher heterogeneity exists in the training data as the medium ASR decreases when \textit{Heterogeneity Index} increases.
Another interesting observation is that the box and whisker become much wider as \textit{Heterogeneity Index} becomes higher, which indicates that the attack effectiveness also becomes less stable when higher heterogeneity presents in training data.  

Backdoor attacks essentially make the model learn the trigger features.
In FL, each client performs its own local training and the local model learns towards reaching the optima of the feature space of that client's local data.
When the training data is more heterogeneous across clients, some features at a client may be more pronounced due to the more skewed local data, i.e., results in overfitting. 
Such more augmented features may suppress backdoor features (e.g., in the extreme case, the backdoor features may become noise compared to the augmented features), and thus make the attack less effective.

\subsection{Redemption 2: An Overlooked Key Factor: Malicious Data Distribution}
\label{sec:redemption2}

\begin{figure*}[h]
    \centering
  \begin{subfigure}{.23\textwidth}
    \centering
    \includegraphics[width=\linewidth]{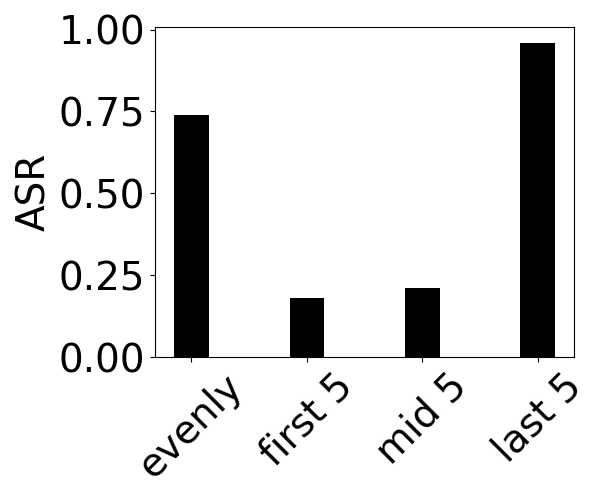}
    \caption{ASR vs. attack timing.}
    \label{fig:curse1_batch_timing}
  \end{subfigure}
  \begin{subfigure}{.5\textwidth}
    \centering
    \includegraphics[width=\linewidth]{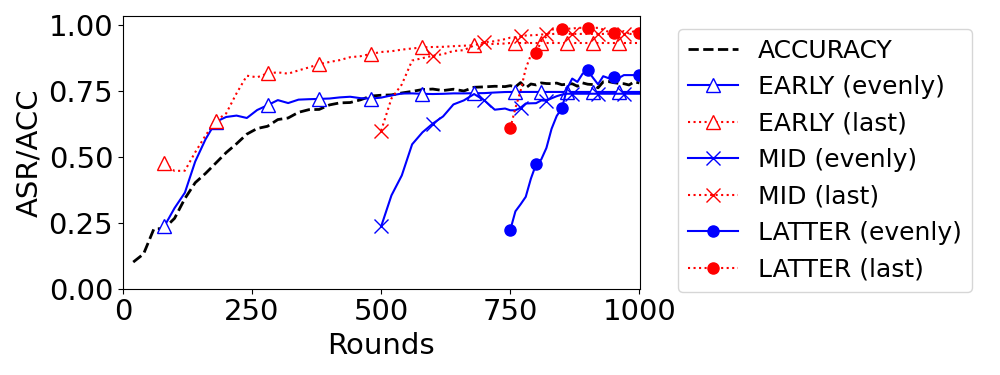}  
    \caption{Global (upper case) and local (lower case) attack timing.}
    \label{fig:curse1_femnist_acc_vs_rounds}
  \end{subfigure}
    \caption{Comparison of different attack timing on FEMNIST.} 
\end{figure*}

\begin{figure*}[h]
  \centering
  \begin{subfigure}{.22\textwidth}
    \centering
    \includegraphics[width=\linewidth]{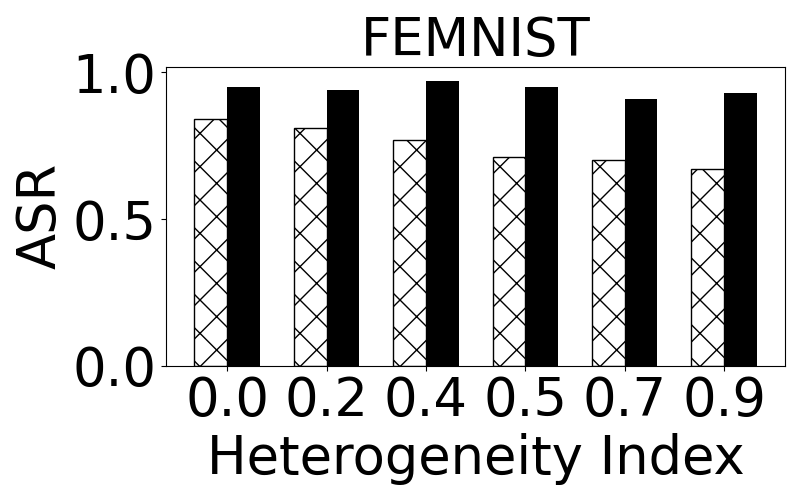}
  \end{subfigure}
  \begin{subfigure}{.22\textwidth}
    \centering
    \includegraphics[width=\linewidth]{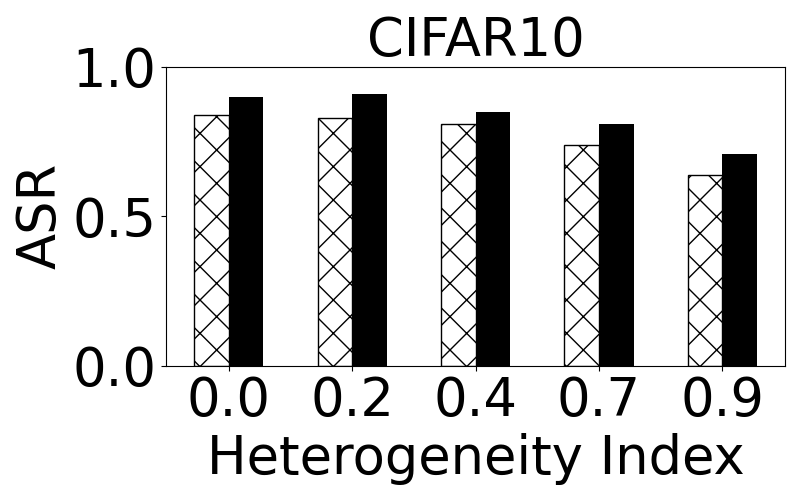}  
  \end{subfigure}
  \begin{subfigure}{.22\textwidth}
    \centering
    \includegraphics[width=\linewidth]{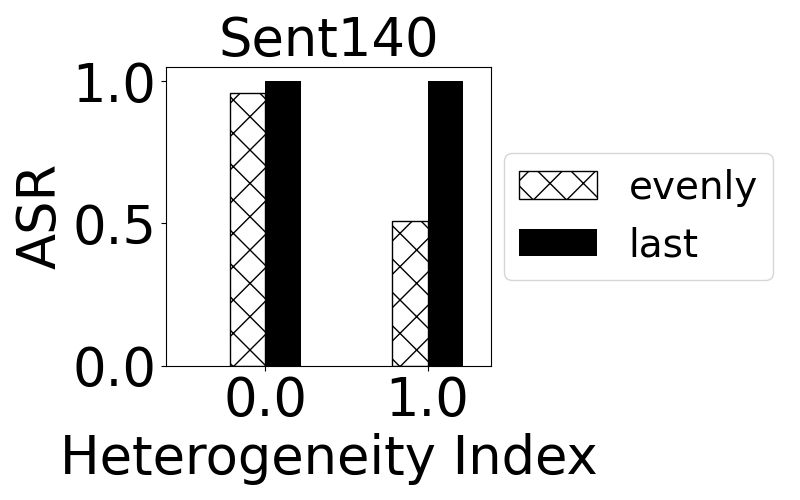}
  \end{subfigure}
  \caption{Comparison between \textit{evenly} vs. \textit{last} batch attack timing under various Heterogeneity Index.} 
  \label{fig:curse1}
\end{figure*}
In Fig.~\ref{fig:redemption1}, even though the trend that data heterogeneity reduces attack effectiveness is clear, from the box-and-whisker plot, we can see that some malicious data distribution is more effective than others.
This indicates that the malicious data distribution can be an important factor in attack effectiveness. Given this has not been studied in the literature, we perform empirical experiments to verify this. 
In this set of experiments, we follow the similar setup as in \textbf{Redemption 1}, except that we fix the \textit{Heterogeneity Index}. Specifically, we use the original training data distribution from LEAF, i.e., \textit{Heterogeneity Index} is 0.2 and 0.0 for FEMNIST and Sent140, respectively.
For CIFAR10, we choose a distribution with \textit{Heterogeneity Index} equal to 0.5.
We report the average ASR for 20 rounds of attack across 25 different malicious data distributions in Fig.~\ref{fig:redemption2}, where each bar represents a unique malicious data distribution. Note that the data distribution of benign clients remains the same. 
The results indeed demonstrate that the attack effectiveness depends on malicious data distribution as the ASR changes significantly when different malicious data distribution is used. 
Such behavior can be explained as the effectiveness of learning backdoor trigger depends on the difference in feature space between training data distribution and malicious data distribution, which we provide further analysis in the \textbf{Curse 3} section. 
This brings a redemption for the robustness of FL as an improper selection of malicious data distribution may result in poor attack effectiveness.

\subsection{Redemption 3: Effective Attack Strategies are More Challenging to Make}
Since malicious data distribution is an important factor in FL backdoor attacks, the natural question is how would it compare to other factors such as the number of attackers and the total number of poisoned datapoints.
To understand this, we conduct experiments by varying the configuration tuple (attack scale, total attack budget, malicious data distribution) and organize the results into a heat map in Fig.~\ref{fig:redemption3}. To make a fair comparison, when we increase the number of attackers, we keep the total number of poisoned datapoints (attack budget) the same and spread evenly across devices. All other parameters are the same as defined in the experimental setup.

The results are quite surprising as there is no clear pattern in the heat maps of all three benchmarks, which is in contrary to the conclusion made by almost all existing work~\cite{bagdasaryan2018backdoor,fung2018mitigating,sun2019can,xie2019dba} that higher \textit{attack scale} and \textit{total attack budget} always lead to more effective attacks.
These counter-intuitive results suggest that the overlooked malicious data distribution is actually a dominant factor in FL backdoor attacks.
Different from homogeneous training data case, where malicious data distribution can be simply configured as IID (the total distribution is a public secret) to maximize the attack effectiveness, malicious data distribution is more difficult to find a reference when training data is heterogeneous.
Unlike the \textit{attack scale} and the \textit{total attack budget}, malicious data distribution is not straightforward to configure, which makes designing effective attack strategies more challenging and the attack effectiveness is thus less predictable.
Because of this, data heterogeneity brings another redemption for the robustness of FL. To demonstrate the observed behaviour is not unique to our chosen attack mechanism, we further evaluated the backdoor attacks proposed in \cite{sun2019can} and \cite{chen2018detecting} and the results (see Appendix) are consistent with Fig.~\ref{fig:redemption3}.

\section{Data Heterogeneity Brings Unseen Curses}

Despite of the redemption brought by data heterogeneity, our further investigations reveal that data heterogeneity can result in several curses for FL backdooring as the attack effectiveness can be significantly boosted by applying proper local attack timing and malicious data distribution, and the backdooring can camouflage itself much easier compared to the homogeneous data case.

\subsection{Curse 1: Local Attack Timing: a New Vulnerability}
 
\begin{figure*}[h!]
  \centering
  \begin{subfigure}{.22\textwidth}
    \centering
    \includegraphics[width=\linewidth]{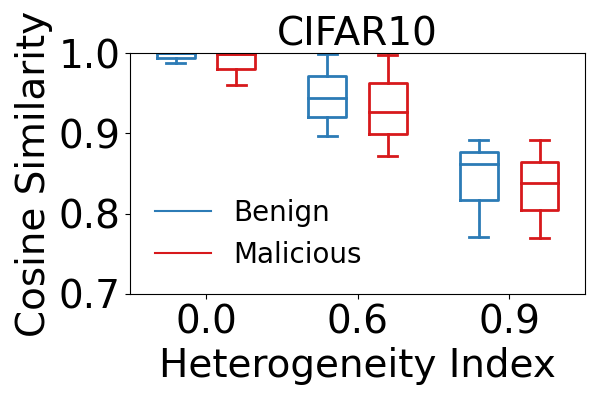}  
  \end{subfigure}
  \begin{subfigure}{.22\textwidth}
    \centering
    \includegraphics[width=\linewidth]{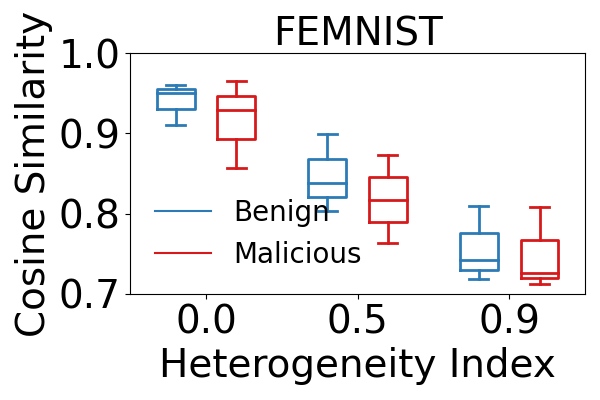}
  \end{subfigure}
  \begin{subfigure}{.22\textwidth}
    \centering
    \includegraphics[width=\linewidth]{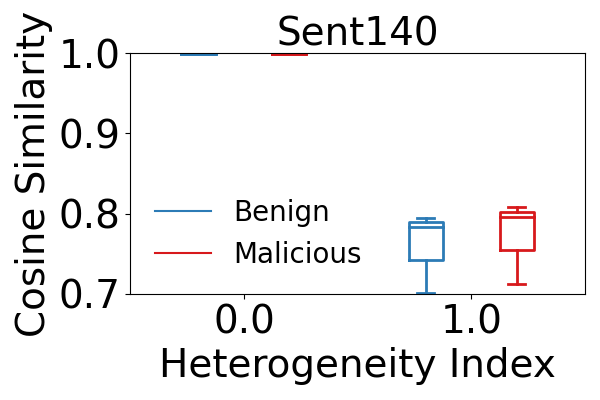}
  \end{subfigure}
  \caption{Cosine Similarity Comparison between benign and malicious clients under different Heterogeneity Index.}
  \label{fig:curse2}
   \vspace{-2mm}
\end{figure*}

\begin{figure*}[h!]
  \centering
  \begin{subfigure}{.24\textwidth}
    \centering
    \includegraphics[width=\linewidth]{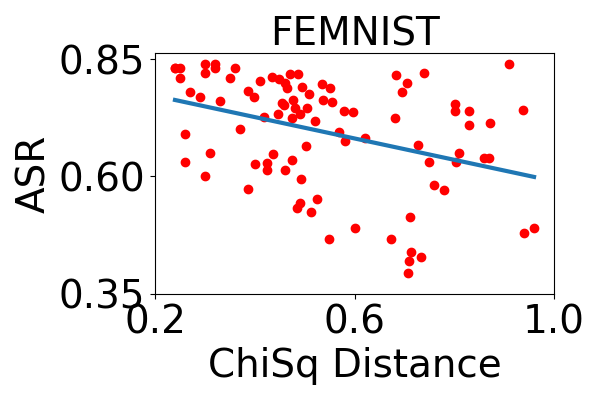}
  \end{subfigure}
  \begin{subfigure}{.24\textwidth}
    \centering
    \includegraphics[width=\linewidth]{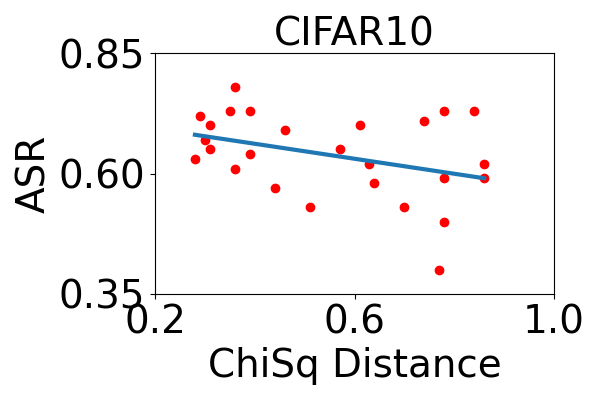}  
  \end{subfigure}
  \begin{subfigure}{.24\textwidth}
    \centering
    \includegraphics[width=\linewidth]{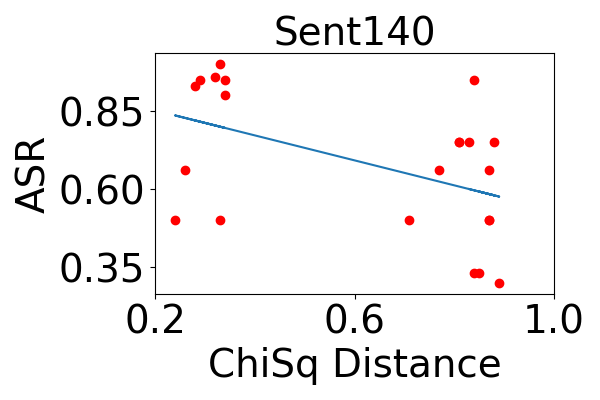}
  \end{subfigure}
      \vspace{-3mm}
  \caption{ASR trend with ChiSq Distance}
  \label{fig:curse3}
      \vspace{-4mm}
\end{figure*}

One important observation is that the local attack timing at each client is important for attack effectiveness, especially with data heterogeneity.
To demonstrate this, we compare four different local attack timing strategies: 1) \textit{evenly} distribute the local attack budget across 10 batches (i.e., the default attack strategy in almost all literature); 2) only attack the \textit{first 5} batches; 3) attack the \textit{middle 5} batches; 4) attack the \textit{last 5} batches.
To make a fair comparison, all the four cases have the same local attack budget, i.e., backdoor 10\% data per batch in \textit{evenly} strategy while backdoor 20\% data per batch for the other three timing strategies.
We use default data heterogeneity of LEAF (i.e., HI=0.2) and all other configures are the same as \textbf{Redemption 1}. The ASR comparison results are presented in Fig~\ref{fig:curse1_batch_timing} and we can see the difference is quite large between different strategies with \textit{last 5} being the highest.
Similar to the reason that data heterogeneity results in less effective attack due to overfitting, here later attack helps backdoor features to be easily overfitted while earlier attack may let the backdoor features easier to be forgotten~\cite{liu2018fine}.  
To understand the behaviors of considering both local and global attack timing, we combine different global attack timing strategies with different local attack timing strategies (\textit{evenly}, \textit{last}).
Note that \textit{last} is attacking only the last batch as we found it performs similar as \textit{last 5} but with 80\% less attack budget 
but with the same \textit{attack scale}.
The comparison results are shown in Fig.~\ref{fig:curse1_femnist_acc_vs_rounds}, where we can see the local attack timing defines the ASR while global attack timing has little impact.
Another important observation is that in \textit{LATTER(last)}, the total attack budget is only 0.2\% of the total training data, one order of magnitude lower than literature~\cite{sun2019can,bagdasaryan2018backdoor,xie2019dba}. Such extremely low budget but highly effective attack makes the local attack timing under data heterogeneity a new vulnerability. We further investigate how data heterogeneity impacts the effects of local attack timing. We perform the same experiments by varying HI and present the results in Fig~\ref{fig:curse1}.

In the \textit{evenly} strategy, as expected, higher heterogeneity results in less attack effectiveness as discussed in \textbf{Redemption 1}. For \textit{last} strategy, it is overall more robust under different heterogeneity and the improvement over \textit{evenly} increases with data heterogeneity. Therefore, the local attack timing can be manipulated by attackers to increase attack effectiveness, especially in high data heterogeneity case.
 
\subsection{Curse 2: Failure of Skewed-Feature Based Defense}

One of the most effective ways to detect FL backdoor attacks is through differentiation between benign features and malicious features (skewed-feature based defense) as they have quite different footprints. 
For instance, cosine similarity can be used to detect anomalous weights~\cite{fung2018mitigating,bagdasaryan2018backdoor}. However, data heterogeneity may increase the weight divergences among the benign clients~\cite{zhao2018federated} thus may make it less distinguishable from malicious clients.
To illustrate this, we use cosine similarity as an example.
Specifically, we compute the cosine similarity of the last dense layer weights of each client against the last dense layer weights of the previous round's global model under different data heterogeneity.
We use the \textit{last} attack timing strategy and the same experiment setup as in \textbf{Redemption 1}.
We use box-and-whisker plot to show the distribution of the cosine similarity values of benign clients and malicious clients respectively in Fig.~\ref{fig:curse2}.
From the results, it is clear that higher data heterogeneity (i.e., higher HI) causes more weights dissimilarity in benign clients (i.e., lower cosine similarity). 
Such high data weights dissimilarity in benign data may be even higher than the dissimilarity of backdoored data, which allows malicious data stealth themselves under the radar of skewed-feature based defense.

\subsection{Curse 3: Malicious Data Distribution as Leverage} 
\label{sec:curse3}
In our experiments from Figure~\ref{fig:redemption1}, we discovered that malicious data distribution is a dominant factor for the attack 
effectiveness and it is more difficult to control compared to attack scale and budget.
With further investigation, we found a simple yet efficient way to generate malicious data distributions that are more effective in attack.
Specifically, we find the distribution distance between malicious data distribution and overall training data distribution is strongly correlated with the attack effectiveness.
We tested a number of divergence metrics such as KL divergence, Jensen-Shannon divergence, Wasserstein distance and B-Distance, and all of them can serve as a good metric here. We use the simple Chi-squared distance (ChiSq or $\chi^2$) as an example for illustration, which is defined as
\begin{equation}\label{chisq_formula}
    \chi^2 =  \tsum_{i=1}^{c}\frac{(O_i - E_i)^2}{E_i},
\end{equation}
\begin{figure}[h]
  \centering
  \begin{subfigure}{.17\textwidth}
    \centering
    \includegraphics[width=\linewidth]{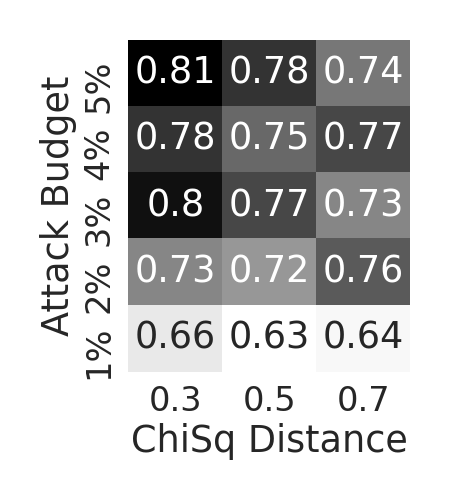}
    \label{fig:redemption3_budget}
  \end{subfigure}
  \begin{subfigure}{.22\textwidth}
    \centering
    \includegraphics[width=\linewidth]{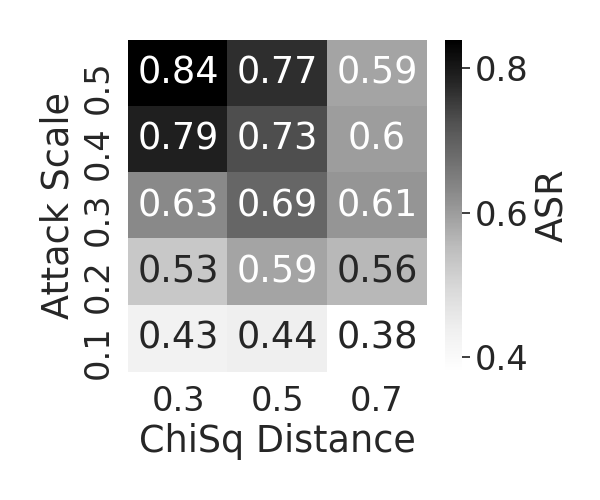}
    \label{fig:redemption3_attacker}
  \end{subfigure}
\vspace{-5mm}
  \caption{ASR comparison between different total attack budget, attack scale, and ChiSq distance.}
  \label{fig:redemption3_2}
\vspace{-3mm}
\end{figure}  
where $E_i$ is the frequency of class $i$ in the training dataset and $O_i$ is frequency of class $i$ in the malicious dataset. 
The smaller the $\chi^2$ value, the more similar the two distributions are.
Intuitively, when drawing a sample from the malicious dataset, it quantifies how close the drawn sample is compared to the training dataset.
To demonstrate the correlation, we do a scatter plot between ASR and ChiSq and perform a linear regression using the scatter points, see Figure~\ref{fig:curse3}. 
The experiments follow the same setup as in \textbf{Redemption 2}.
The regression curve demonstrates a good correlation between ASR and ChiSq and the points are more clustered when ChiSq distance is smaller.
To verify this, we perform experiments by varying the configuration tuples (total attack budget, ChiSq) and (attack scale, ChiSq) respectively and organize the results into heat maps, see Fig.~\ref{fig:redemption3_2}. The results show that overall lower ChiSq attack achieves better ASR and can even outperform attacks with higher budget but also higher ChiSq. 
Although these results are ``expected'', it is contrary to the findings in Fig.~\ref{fig:redemption3}, which indicates that the existing works on robustness of FL have not been fully evaluated on stronger attacks.
\section{Defending the Curses Brought by Data Heterogeneity}
In this section, we discuss the challenges and potential directions of defending the curses brought by data heterogeneity in FL backdoor attacks. 

\begin{table}[h!]
\label{table:def1}
\centering
\caption{Attack Success Rate comparison between without and with the proposed active defense.}\label{setup_table}
\scalebox{0.9}{
\begin{tabular}{|c|c|c|}
\hline
\bf Dataset  &\bf ASR w/o Defense & \bf ASR w/ Defense
\\ \hline
CIFAR10 & 0.76 & 0.26 \\ \hline
FEMNIST & 0.96 & 0.21 \\ \hline
Sent140 & 1.0 & 0.36 \\ \hline
\end{tabular}}
\vspace{-3mm}
\end{table}

\textbf{Defending Curse 1: Cut the Short Path of Overfitting.} Backdooring the last batch of a malicious client results in overfitting of the local model on triggered data samples. 
Accumulating the overfitted model weights of malicious clients to the global model may lead to high ASR. 
To defend against such a strategy, evading the overfitted weight updates during the aggregation process is critical. 
There is a rich line of work for addressing this problem in traditional ML~\cite{shen2016auror,wang2019neural,liu2018fine}, but all of them require knowledge from the training data, which is infeasible in FL due to privacy requirement.
Therefore, we propose an active defense mechanism in which the aggregator assumes all clients are malicious. 
The aggregator maintains a global (but small) IID dataset to train the updated weights of all the participating clients before aggregation.
The overfitting due to backdoor triggers is thus minimized and the model becomes more generalizable.
This mechanism is inspired by a previous paper~\cite{zhao2018federated}, where the goal is to increase task accuracy while we focus on mitigating attack effectiveness.
The evaluation results are presented in Table~\ref{setup_table}, where we use an IID dataset with a size equal to 10\% of the total dataset on the aggregator.
The results show ASR is significantly reduced after applying this defense.
The limitation of this method is that if secure aggregation is used, it may be difficult to train individual client on the IID dataset.

\textbf{Defending Curse 2: An Overfitting Mitigating Mechanism for Client Selection.}
Given skewed-feature based defense is difficult to distinguish whether the overfitting is from data heterogeneity or malicious attack, we suggest diversifying the selection of clients so that even if the local model is overfitted by backdoor triggers, the overfitted local model weights have less chance to be accumulated to the global model.  
We implemented a scheduling policy as proof of concept to avoid selecting the same client in nearby rounds (e.g., a client needs to wait at least 20 rounds to be selected again) so that the malicious clients are spreading out further away, which allows FL to forget backdoors easier over time. 
The results show that with the help of this defend policy, ASR decreases across every heterogeneity level and none of them achieves over 23\% ASR. 
We also plan to investigate more complex detection methods such as using activation clustering~\cite{chen2018detecting}, spectral signatures~\cite{tran2018spectral}, and gradient shaping~\cite{hong2020effectiveness} in our future work and potentially combine them with the client selection mechanism.
\textbf{Defending Curse 3: Protect the Training Data Distribution.}
As observed in \textbf{Curse 3}, attackers can design an efficient attack by generating a similar malicious data distribution as the global data.
Existing works that change or augment training data still preserve its distribution and thus difficult to be employed here~\cite{shen2016auror,liu2018fine,tran2018spectral,wang2019neural}.
To defend such attack strategies, we need to avoid revealing the global data distribution. We also set up a simple experiment where we simulate faking the actual global data distribution, and the malicious clients end up building their attack based on a distribution that has a high Chi-Squared value (e.g., about 0.8 in our experiments) compared to the real global distribution. With this defending strategy, the ASRs are much lower -- on average 0.46 (reduced from on average 0.8). 
When this is not possible, we can try to mislead the attackers to believe a wrong global data distribution. We can also try to disrupt the global data distribution, such as having extra data reserved at the aggregator (similar to the proposal in \textit{Defending Curse 1}), or through GAN like data anonymization~\cite{hukkelaas2019deepprivacy}, which can be used to design a more robust aggregation method.

\section{Conclusion}
In this paper, we perform extensive empirical experiments to quantify and understand the impact brought by data heterogeneity in backdoor attacks of federated learning. We identified several redemptions and curses, and proposed some potential remedy strategies. The results show that depending on the extent of data heterogeneity the impacts of backdooring can vary significantly. The lessons learned here offer new insights for designing defenses for Federated Learning.
\section{Acknowledgements}
This work is supported in part by the following grants: National Science Foundation CCF-1756013 and IIS-1838024 (with resources from AWS as part of the NSF BIGDATA program). We thank the anonymous reviewers for their insightful comments and suggestions.
\bibliography{ref}
\newpage
\section{Appendix}
\begin{figure*}[th!]
  \centering
  \centering
  \begin{subfigure}{.20\textwidth}
    \centering
    \includegraphics[width=\linewidth]{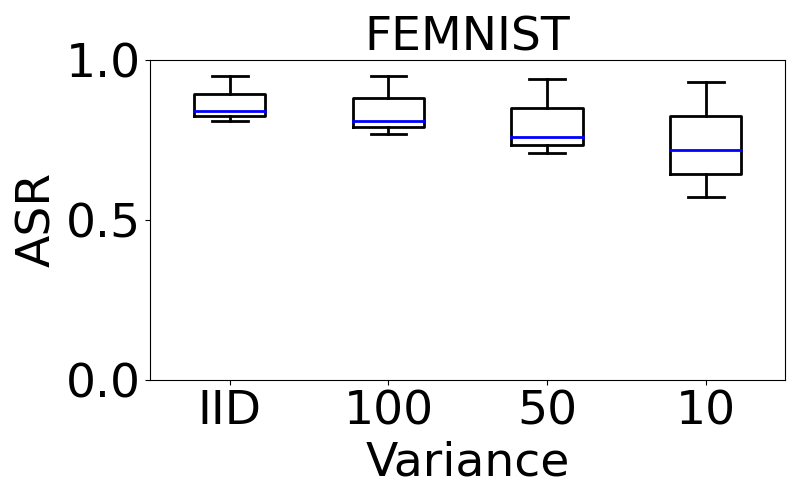}
  \end{subfigure}
  \begin{subfigure}{.20\textwidth}
    \centering
    \includegraphics[width=\linewidth]{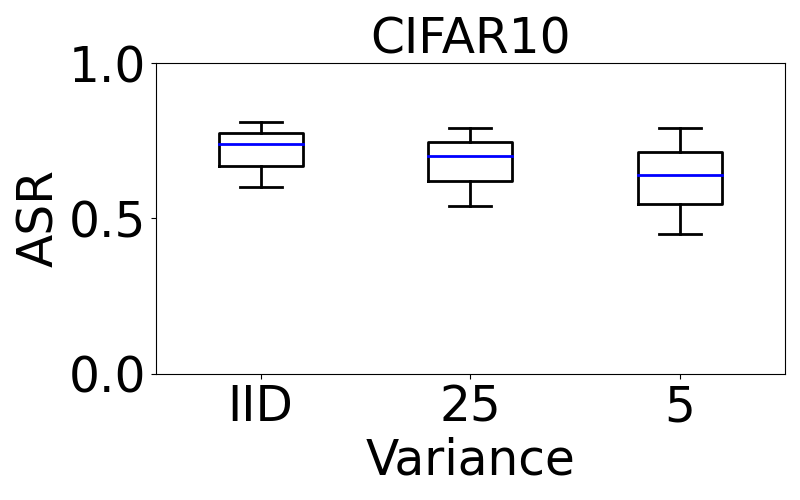}
  \end{subfigure}
  \begin{subfigure}{.20\textwidth}
    \centering
    \includegraphics[width=\linewidth]{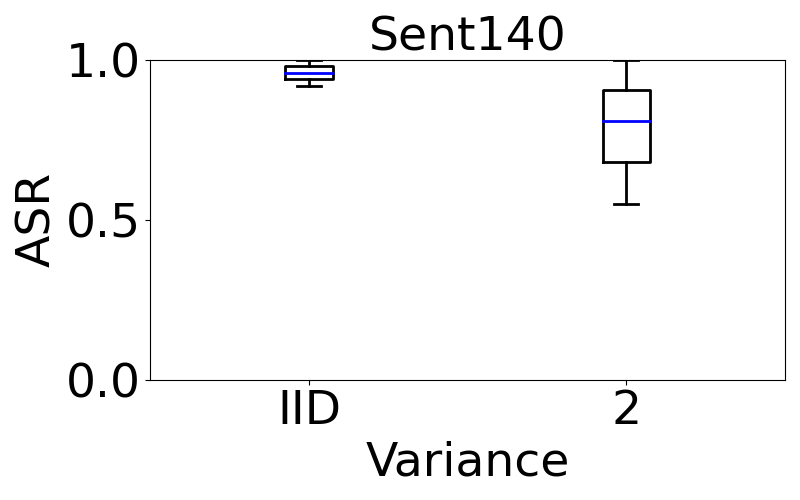}
  \end{subfigure}
  \vspace{-3mm}
  \caption{Attack Success Rate (ASR) vs. Data Heterogeneity simulated by Gaussian Sampling (lower variance represents higher heterogeneity).} 
  \label{fig:appendix_1}
  \vspace{-3mm}
\end{figure*}
\begin{figure*}[h!]
  \centering
  \centering
  \begin{subfigure}{.20\textwidth}
    \centering
    \includegraphics[width=\linewidth]{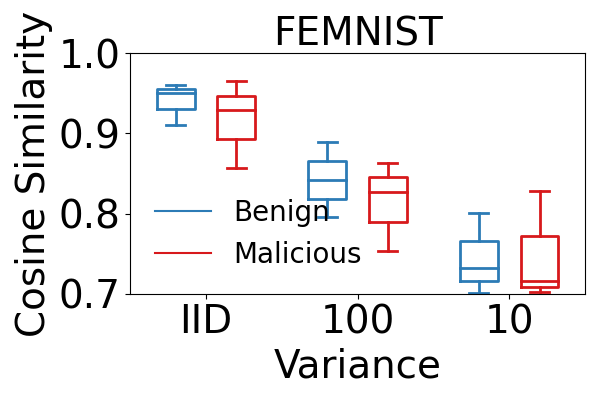}
  \end{subfigure}
  \begin{subfigure}{.20\textwidth}
    \centering
    \includegraphics[width=\linewidth]{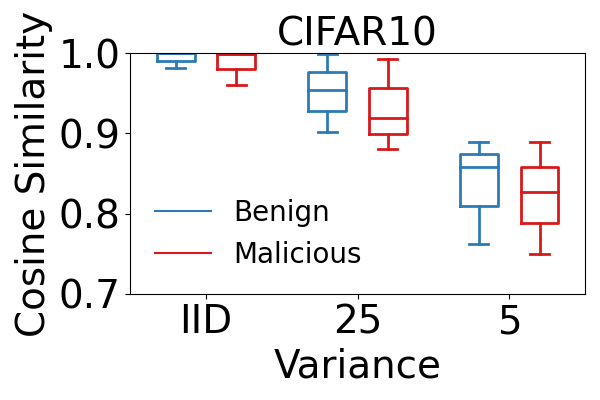}
  \end{subfigure}
  \begin{subfigure}{.20\textwidth}
    \centering
    \includegraphics[width=\linewidth]{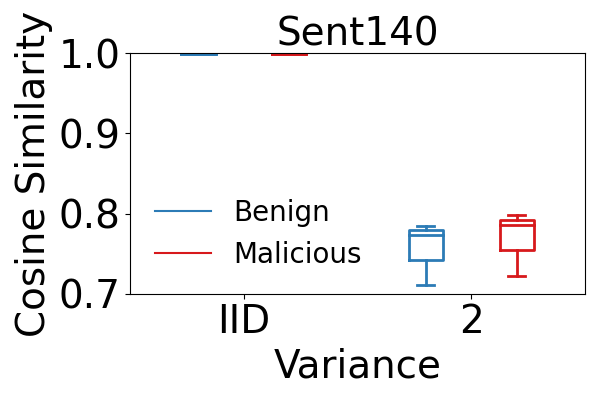}
  \end{subfigure}
  \vspace{-3mm}
  \caption{Cosine Similarities comparison between benign and malicious clients under different Data Heterogeneity simulated by Gaussian Sampling (lower variance represents higher heterogeneity).}
  \label{fig:appendix_2}
  \vspace{-3mm}
\end{figure*}
\subsection{Data Heterogeneity by Gaussian Sampling}

Almost all existing works in federated learning simulate data heterogeneity 
by limiting the number of classes available in each client~\cite{zhao2018federated, chai2020tifl, bonawitz2017practical, li2019convergence, sattler2019robust}. 
We followed existing works when conducting experiments in the main draft.
To evaluate whether our findings are robust to different data heterogeneity, here we provide another way to simulate the data heterogeneity by using Gaussian sampling~\cite{bhagoji2019analyzing}. We employ Gaussian Sampling to sample data from the total dataset for creating dataset for each client. The heterogeneity of data can be controlled by tuning the variance of the Gaussian distribution used for sampling (in Gaussian Sampling, a higher variance represents a wider distribution of data sampling), which correlates to the diversity of the features in the sampled datasets that determines the data heterogeneity. In other words, a higher variance represents the case that we select a more diverse set of data points from the total dataset.
We generate the same Attack Success Rate and Cosine Similarities plots as in the main draft (i.e., Figure 2 and Figure 7 in the main draft) and shown in Figure~\ref{fig:appendix_1} and Figure~\ref{fig:appendix_2}. 
We can see the observations obtained in the main draft are consistent with the results present here, which verifies that our findings hold under different ways of simulating data heterogeneity.

\subsection{Weight Scaling Factor Analysis}

\begin{figure*}[h!]
  \centering
  \centering
  \begin{subfigure}{.20\textwidth}
    \centering
    \includegraphics[width=\linewidth]{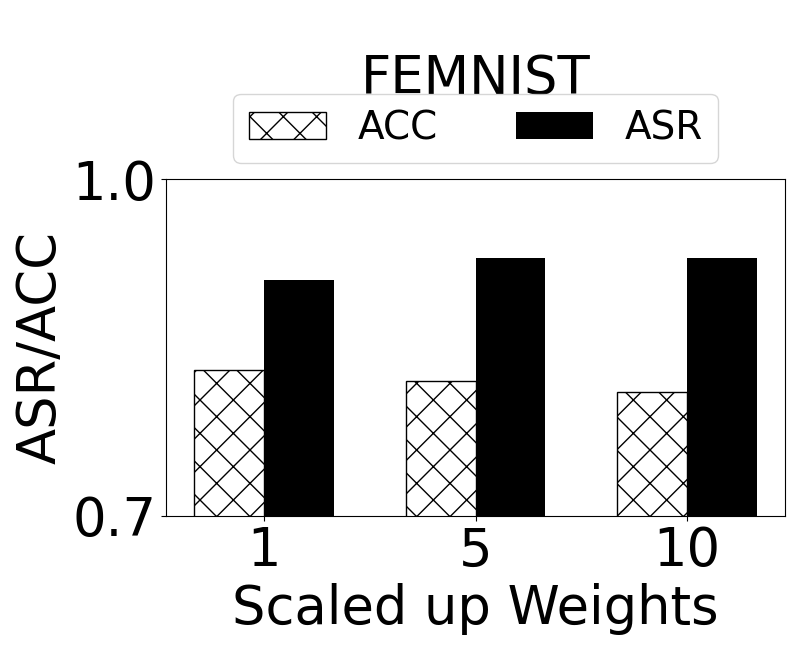}
  \end{subfigure}
  \begin{subfigure}{.20\textwidth}
    \centering
    \includegraphics[width=\linewidth]{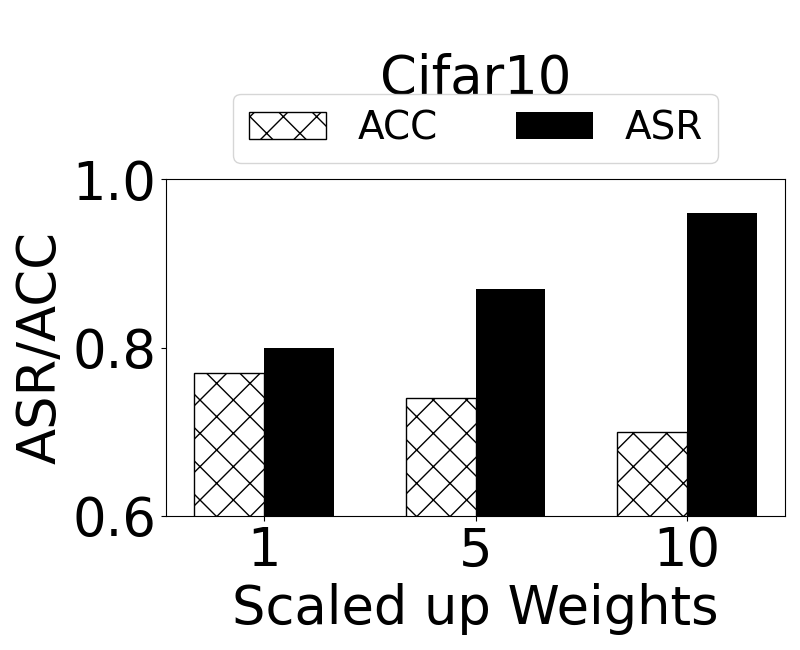}
  \end{subfigure}
  \begin{subfigure}{.20\textwidth}
    \centering
    \includegraphics[width=\linewidth]{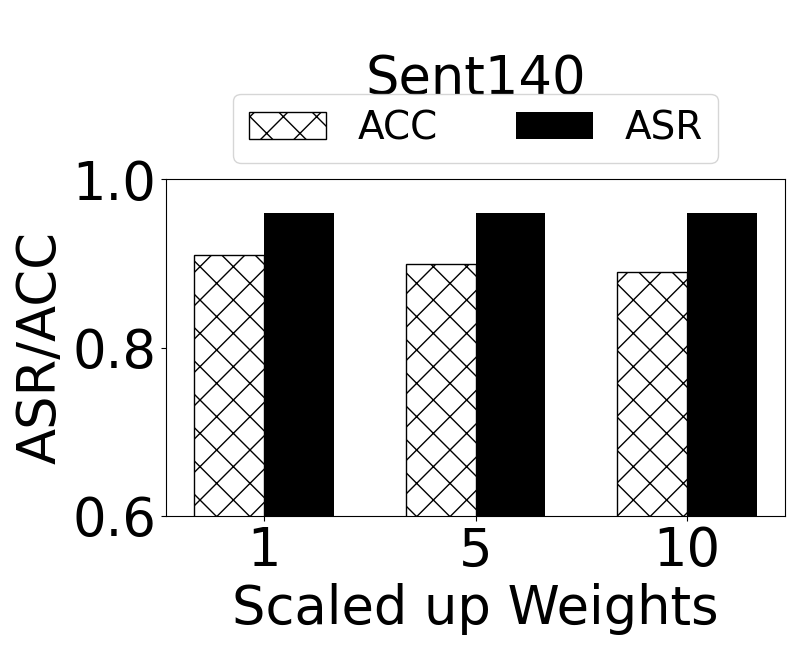}
  \end{subfigure}
  \vspace{-3mm}
  \caption{Attack success rate (ASR) and accuracy (ACC) comparison under different weight scaling factor (1, 5, 10).} 
  \label{fig:appendix_3}
  \vspace{-3mm}
\end{figure*}

\textit{Model replacement attacks} (aka model poisoning attacks) attempt to replace the benign model with a malicious model, which is what backdooring on local devices aims to achieve. The malicious clients train backdoor into their local models and then send the weights to the server in an attempt to make the aggregation algorithm replace the global model with the backdoored model. 
If the malicious weights during weights aggregation are pronounced enough, the malicious weights can overwhelm the aggregation process to cause model replacement attacks.
As pointed out in~\cite{bagdasaryan2018backdoor}, data poisoning attacks in federated learning are in fact subsumed by model replacement attacks.
To demonstrate this, we run experiments by scaling up the weights of the models by a factor of 5 and 10 respectively and plot the corresponding attack success rate (ASR) and accuracy (ACC) in  Figure~\ref{fig:appendix_3}. 
The results show that with scaled up weights, the attack success rate is only slightly better but the model accuracy is decreased. This suggests the findings of backdooring attack in this paper can be generalized to model replacement attack. 
However, scaling up the weights in practice is difficult to achieve due to the privacy protection mechanism such as~\cite{abadi2016deep, bonawitz2017practical}. 
In addition, scaled up weights can be detected as outliers compared to weights of benign clients~\cite{sun2019can,bagdasaryan2018backdoor,xie2019dba}.
Therefore, in the main draft, we focus on non-scaled weights case (i.e., weight scaling factor is 1).

\subsection{Different Attack Strategies}

\begin{figure}[h!]
  \centering
  \begin{subfigure}{.20\textwidth}
    \centering
    \includegraphics[width=\linewidth]{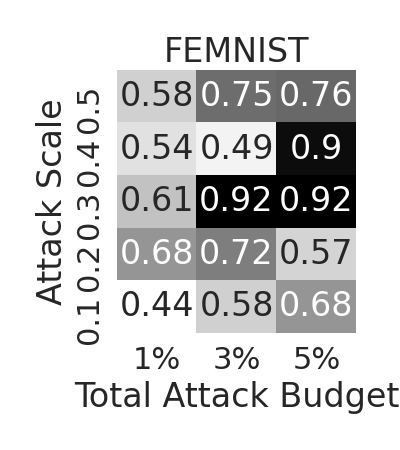}
  \vspace{-7mm}
    \caption{Attack 1}
  \end{subfigure}
  \begin{subfigure}{.20\textwidth}
    \centering
    \includegraphics[width=\linewidth]{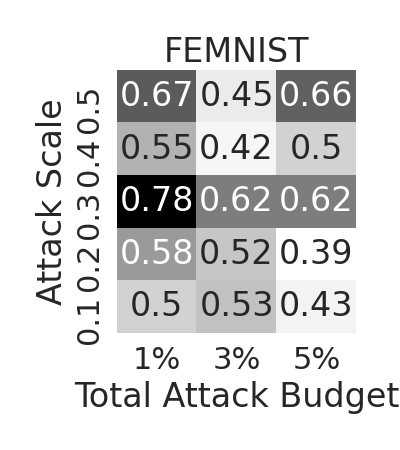}
  \vspace{-7mm}
    \caption{Attack 2}
  \end{subfigure}
  \begin{subfigure}{.20\textwidth}
    \centering
    \includegraphics[width=\linewidth]{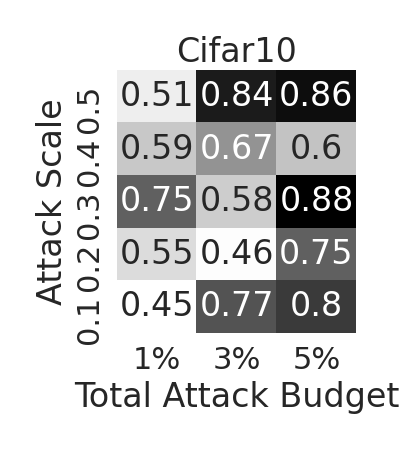}
  \vspace{-6mm}
    \caption{Attack 1}
  \end{subfigure}
  \begin{subfigure}{.20\textwidth}
    \centering
    \includegraphics[width=\linewidth]{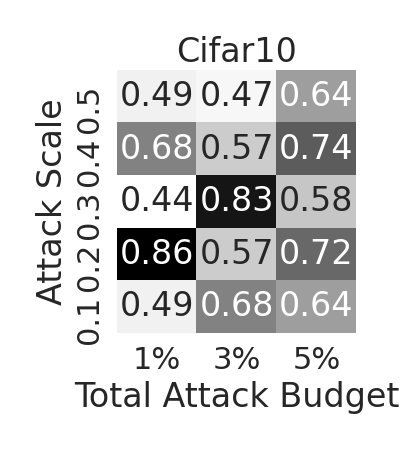}
  \vspace{-6mm}
    \caption{Attack 2}
  \end{subfigure}
  \vspace{-3mm}
  \caption{Attack Success Rate (ASR) scalability in terms of attack scale and total attack budget when using Attack 1 and Attack 2.}
  \label{fig:appendix_4}
  \vspace{-3mm}
\end{figure}

We use the attack strategy proposed in~\cite{xie2019dba} for our analysis in the main draft as it claims more efficient attack than other existing works. To ensure our findings are robust to different attacks, we also conduct the experiments using the attack strategies proposed in~\cite{sun2019can} and~\cite{chen2018detecting} (labeled Attack 1 and Attack 2 respectively).
Specifically, we run the same experiments as shown in Figure 4 of the main draft to evaluate the Attack Success Rate (ASR) scalability in terms of attack scale and total attack budget, see Figure~\ref{fig:appendix_4}. It is worth noting that the results of Sent140 is not included here because both the strategies focus on image-based applications. 
We get the same counter-intuitive results when using the attack strategy proposed in~\cite{xie2019dba}: there is no clear pattern in the heat maps, which is in contrary to the conclusion made by almost all existing work~\cite{bagdasaryan2018backdoor,fung2018mitigating,sun2019can,xie2019dba} that higher \textit{attack scale} and \textit{total attack budget} always leads to more effective attacks.

\subsection{Evaluation on Defense Strategies}
For defense, we proposed several strategies in the main draft by taking the data heterogeneity into account, which is overlooked by existing defense methods. 
For strategies \textbf{Defending Curse 2} proposed in the main draft, one of the defense method is to diversify the selection of clients so that even if the local model is overfitted by backdoor triggers, the overfitted local model weights have less chance to be accumulated to the global model. To verify the effectiveness of such strategy, we implement a uniform random selection policy with a \textit{selection separation factor} defined as the minimum number of rounds that a client can be selected.
We present the results in Table~\ref{table:appendix_5} where we show the Attack Success Rates under different \textit{selection separation factor} values. 
We can see when the factor is increasing, the Attack Success Rate drops significantly. 
Therefore, we consider spacing out client selection is a promising defense strategy for defending Curse 2.


\begin{table}[h!]
\centering
\caption{Defending Curse 2: spreading out client selection over rounds. The value reported in the table is the Attack Success Rate.} 
\begin{tabular}{c|c|c|c|}
\cline{2-4}
\multicolumn{1}{l|}{}                  & \multicolumn{3}{c|}{\textbf{Selection Separation Factor}} \\ \cline{2-4} 
\multicolumn{1}{l|}{}                  & \textbf{10}       & \textbf{20}       & \textbf{50} \\ \hline
\multicolumn{1}{|c|}{\textbf{FEMNIST}} &  \hspace{1.6mm} 0.87    \hspace{1.5mm}          & \hspace{1.5mm} 0.16    \hspace{1.5mm}     & 0.14   \\ \hline
\multicolumn{1}{|c|}{\textbf{Cifar10}} & 0.83              & 0.23              & 0.11  \\ \hline
\multicolumn{1}{|c|}{\textbf{Sent140}} & 1.0               & 0.2               & 0.2  \\ \hline
\end{tabular}
\label{table:appendix_5}
\end{table}

For strategies \textbf{Defending Curse 3} proposed in the main draft, one approach is to mislead the attackers to believe a false global data distribution.
To verify this idea, we generate false global data distributions based on the Chi-squared (ChiSq) distance of the true global data distribution and on purposely disclosure this false distribution information to the attackers. 
In Curse 3, attackers can leverage the (true) global data distribution to generate highly effective attacks. 
However, when they use the false global data distribution to generate attacks, the attack success rate is significantly dropped and the larger the ChiSq distance between false and true global data distribution, the larger drop in attack success rate, see Table~\ref{table:appendix_6}.
Therefore, if we can mislead attackers to believe a false global data distribution, we can defense well backdooring attacks.

\begin{table}[h!]
\centering
\caption{Defending Curse 3: mislead attackers to believe a false global data distribution. The value reported in the table is the Attack Success Rate. The ChiSq distance is computed between false and true global data distribution.} 
\begin{tabular}{c|c|c|c|c|}
\cline{2-5}
                                       & \multicolumn{4}{c|}{\textbf{ChiSq Distance}}              \\ \cline{2-5} 
                                       & \textbf{0.0} & \textbf{0.5} & \textbf{0.7} & \textbf{0.9} \\ \hline
\multicolumn{1}{|c|}{\textbf{FEMNIST}} & 0.91         & 0.45         & 0.33         & 0.24         \\ \hline
\multicolumn{1}{|c|}{\textbf{Cifar10}} & 0.84         & 0.33         & 0.23         & 0.18         \\ \hline
\multicolumn{1}{|c|}{\textbf{Sent140}} & 1.0          & 0.2          & 0.1          & 0.1          \\ \hline
\end{tabular}
\label{table:appendix_6}
\end{table}

\end{document}